# Estimating Phoneme Class Conditional Probabilities from Raw Speech Signal using Convolutional Neural Networks

*Dimitri Palaz*[1,2], *Ronan Collobert*[1], *Mathew Magimai.-Doss*[1]

[1]Idiap Research Institute, Martigny, Switzerland
[2]Ecole Polytechnique Fédérale de Lausanne (EPFL), Lausanne, Switzerland
{dimitri.palaz, ronan.collobert, mathew}@idiap.ch

## Abstract

In hybrid hidden Markov model/artificial neural networks (HMM/ANN) automatic speech recognition (ASR) system, the phoneme class conditional probabilities are estimated by first extracting acoustic features from the speech signal based on prior knowledge such as, speech perception or/and speech production knowledge, and, then modeling the acoustic features with an ANN. Recent advances in machine learning techniques, more specifically in the field of image processing and text processing, have shown that such divide and conquer strategy (i.e., separating feature extraction and modeling steps) may not be necessary. Motivated from these studies, in the framework of convolutional neural networks (CNNs), this paper investigates a novel approach, where the input to the ANN is raw speech signal and the output is phoneme class conditional probability estimates. On TIMIT phoneme recognition task, we study different ANN architectures to show the benefit of CNNs and compare the proposed approach against conventional approach where, spectral-based feature MFCC is extracted and modeled by a multilayer perceptron. Our studies show that the proposed approach can yield comparable or better phoneme recognition performance when compared to the conventional approach. It indicates that CNNs can learn features relevant for phoneme classification automatically from the raw speech signal.

## 1. Introduction

Hidden Markov model (HMM) based automatic speech recognition (ASR) system, similar to conventional pattern recognition system, breaks the problem into several sub-tasks: feature extraction, modeling and decision making, and optimizes them in independent manner. For instance, acoustic features such as, mel frequency cepstral coefficients (MFCC), perceptual linear prediction (PLP) cepstral coefficients, linear prediction cepstral coefficients are extracted based on prior knowledge about speech perception and/or speech production. These features are then usually modeled by either Gaussian mixture models (GMM) or artificial neural networks (ANNs) to estimate state emission distribution. This step is often referred to as acoustic modeling. The decision making, i.e. recognition, step integrates the acoustic model, lexical knowledge and language model/syntactical constraints (again estimated independently on text data) to decode the test utterance.

In recent years, in the field of computer vision [1] and text processing [2] studies on sequence recognition problems similar to ASR have shown that such divide and conquer strategy may not be necessary. More precisely, these studies have shown that it is possible to build end-to-end systems (fed with raw input data) by using architectures composed of many layers, where each layer *learns* features (i.e. abstract representations), that are relevant to the problem of interest.

Inspired from these studies, the present paper, as a first modest step, investigates estimation of phoneme class conditional probabilities from raw speech signal using convolutional neural networks[1] (CNN) [4] for phoneme sequence recognition. In the framework of hybrid HMM/ANN system, we compare the proposed approach with the conventional approach of extracting spectral-based acoustic feature extraction and then modeling them by ANN. In addition, we also propose a discriminative decoding algorithm based on a simple conditional random field (CRF). Experimental studies conducted on TIMIT corpus show that (a) the proposed approach can yield a phoneme recognition system that is similar to or better than the system based on conventional approach and (b) CRF-based decoding yields better performance than conventional joint likelihood based decoding.

The remainder of the paper is organized as follows. Section 2 presents a brief survey of related literature. Section 3 presents the architecture of the proposed system. Section 4 presents the experimental setup and Section 5 presents the results. Section 6 presents an analysis, Section 7 provides a discussion and Section 8 concludes the paper.

## 2. Related Work

Despite the success of spectral-based acoustic features, there has been interest in modeling raw speech signal for speech recognition. In one of the earliest work, Poritz proposed an approach where the speech signal is modeled by a linear prediction HMM [5]. This work was later revisited as switching autoregressive HMM [6], and more recently in the framework of switching linear dynamical systems [7]. Experiments on isolated word/digit recognition task have shown that these approaches can yield performance comparable to standard cepstral-based HMM system in clean conditions, and better performance under noisy conditions [7]. In [8], an approach to model raw speech signal was proposed. In this

---
This work was partly supported by the HASLER foundation through the grant "Universal Spoken Term Detection with Deep Learning" (DeepSTD) and by the Swiss NSF through the Swiss National Center of Competence in Research (NCCR) on Interactive Multimodal Information Management (www.im2.ch).

---
[1]In speech literature, CNN is referred to as time-delay neural network [3].

approach, the signal statistical characteristics are modeled as the output of a filter excited by a Gaussian source. The potential of the approach was demonstrated on classification of speaker-dependent discrete utterances consisting of 18 highly confusable stop consonant-vowel syllables. More recently, combination of raw speech and cepstral features in the framework of support vector machine has been investigated for noisy phoneme classification [9].

In recent years, there has been growing interests in using short-term spectrum as features, mainly in the framework of artificial neural networks. These "intermediate" representations (standing between raw signal and "classical" features such as cepstral-based features) have been successfully used in speech recognition applications [10, 11, 12, 13].

## 3. Proposed system

CNNs are a particular kind of artificial neural network which performs a series of convolutions over the input signal. They learn convolution filters in a end-to-end manner from raw data, alleviating the problem of designing/choosing the right features for a particular task of interest. CNN-based systems have been shown to lead to state-of-the-art performance on image [14, 15] or text [2] problems. In this paper, we show that the *convolutional* aspect of CNNs make them particularly suitable for handling temporal signals such as raw speech.

The proposed system is composed of two parts: the estimation of the phoneme class conditional probabilities and the decoding of the sequence. The first part is performed by a CNN, which takes raw speech signal as input. For second part, a simple CRF will be used to decode the sequence.

### 3.1. Convolutional Neural Network

The network is given a window of raw input signal and computes the conditional probability $p(i|x)$ for each phoneme class $i$. One class is then attributed to an example by computing $\mathrm{argmax}(p(i|x))$. These type of network architectures are composed of several filter extraction stages, followed by a classification stage. A filter extraction stage involves a convolutional layer, followed by a temporal pooling layer and an non-linearity ($\tanh()$). Our optimal architecture included 3 stages of filter extraction (see Figure 1). Signal coming out of these filter stages are fed to a classification stage, which in our case was a one-hidden layer MLP. The last layer is a *softmax* layer, which computes the conditional probability.

#### 3.1.1. Convolutional layer

While "classical" linear layers in standard MLPs accept a fixed-size input vector, a convolution layer is assumed to be fed with a sequence of $T$ vectors/frames: $X = \{x^1 \ x^2 \ \ldots \ x^T\}$. A convolutional layer applies the same linear transformation over each successive (or interspaced by $dW$ frames) windows of $kW$ frames. E.g, the transformation at frame $t$ is formally written as:

$$M \begin{pmatrix} x^{t-(kW-1)/2} \\ \vdots \\ x^{t+(kW-1)/2} \end{pmatrix}, \quad (1)$$

where $M$ is a $d_{out} \times d_{in}$ matrix of parameters. In other words, $d_{out}$ filters (rows of the matrix M) are applied to the input sequence. An illustration is provided in Figure 2.

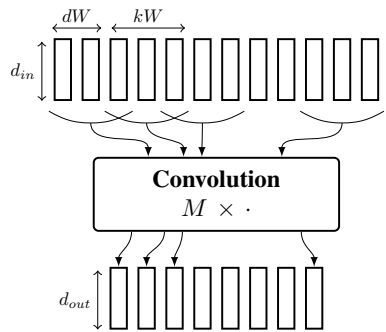

Figure 2: *Illustration of a convolutional layer. $d_{in}$ and $d_{out}$ are the dimension of the input and output frames. $kW$ is the kernel width (here $kW = 3$) and $dW$ is the shift between two linear applications (here, $dW = 2$).*

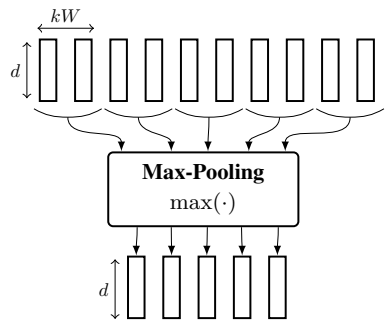

Figure 3: *Illustration of max-pooling layer. $kW$ is the number of frame taken for each $\max$ operation and $d$ represents the dimension of input/output frames (which are equal).*

#### 3.1.2. Max-pooling layer

These kind of layers perform local temporal $\max$ operations over an input sequence, as shown in Figure 3. More formally, the transformation at frame $t$ is written as:

$$\max_{t-(kW-1)/2 \leq s \leq t+(kW-1)/2} f_i^s \quad \forall i \quad (2)$$

These layers increase the robustness of the network to slight temporal distortions in the input.

#### 3.1.3. SoftMax layer

The $Softmax$ [16] layer interprets network output scores $f_i(x)$ as conditional probabilities, for each class label $i$:

$$p(i|x) = \frac{e^{f_i(x)}}{\sum_j e^{f_j(x)}} \quad (3)$$

#### 3.1.4. Network training

The network parameters $\theta$ are learned by maximizing the log-likelihood $L$, given by:

$$L(M_1, ..., M_L, \theta) = \sum_{n=1}^{N} \log(p(i_n|x_n, \theta)) \quad (4)$$

for each input $x$ and label $i$, over the whole training set, with respect to the parameters of each layer $M_l$. Defining the `logadd`

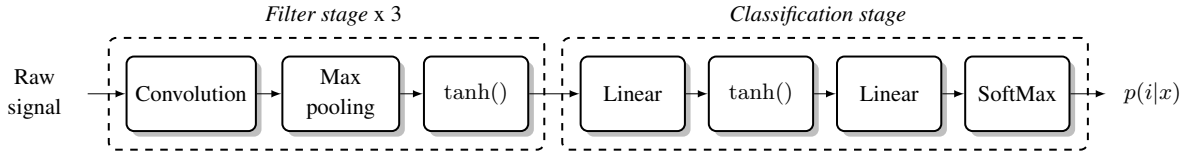

Figure 1: *Convolutional Neural Network. Several stages of convolution/pooling/tanh might be considered. Our network included 3 stages.*

operation as: $\text{logadd}_i(z_i) = \log(\sum_i e^{z_i})$, the likelihood $L$ can be expressed as:

$$L = \log(p(i|x)) = f_i(x) - \underset{j}{\text{logadd}}(f_j(x)) \quad (5)$$

where $f_i(x)$ described the network score of input $x$ and class $i$. Maximizing this likelihood is performed using the stochastic gradient ascent algorithm [17].

### 3.2. Decoder

We consider a very simple version of CRFs, where we define a graph with nodes for each frame in the input sequence, and each label. This CRF allows to discriminatively train a simple duration model over our network output scores. Transition scores are assigned to edges between phonemes, and network output scores are assigned to nodes. Given an input data sequence $x$ and a label path on the graph $y$, a score for the path can be defined:

$$s(x, y) = \sum_{t=1}^{T} \left( f_{y_t}(x_t) + A_{y_t, y_{t-1}} \right) \quad (6)$$

where $A$ is a matrix describing transitions between labels and $f_{y_t}(x_t)$ the network score of input $x$ for class $y$ at time $t$. *Path* scores are interpreted as conditional probabilities, by applying a softmax (see Section 3.1.3) over all possible paths. The CRF transitions scores are then trained by maximizing the likelihood over the training data, with a gradient ascent.

## 4. Experimental Setup

In this section we present the setup used for the experiments, as well as the different features and the decoding algorithms.

### 4.1. TIMIT Corpus

The TIMIT acoustic-phonetic corpus consists of 3,696 training utterances (sampled at 16kHz) from 462 speakers, excluding the SA sentences. The cross-validation set consists of 400 utterances from 50 speakers. The core test set was used to report the results. It contains 192 utterances from 24 speakers, excluding the validation set. The 61 hand labeled phonetic symbols are mapped to 39 phonemes with an additional garbage class, as presented in [18].

### 4.2. Features

Raw features are simply composed of a window of the speech signal (hence $d_{in} = 1$, for the first convolutional layer as shown in Figure 1). The window is normalized such that it has zero mean and unit variance.

We also performed several experiments, with MFCC as input features. They were computed (with HTK [19]) using a 25 ms Hamming window on the speech signal, with a shift of 10 ms. The signal is represented using 13th-order coefficients along with their first and second derivatives, computed on a 9 frames context ($d_{in} = 39$ for the first convolutional layer).

### 4.3. Network hyper-parameters

The hyper-parameters of the network are: the input window size, corresponding to the context taken along with each example, the kernel width $kW$ and shift $dW$ of the convolutions, the number of filters $d_{out}$, the width of the hidden layer and the pooling width. They were tuned by early-stopping on the cross-validation set. Ranges which were considered for the grid search are reported in Table 1. It is worth mentioning that for a given input window size over the raw signal, the size of the output of the filter extraction stage will strongly depend on the number of max-pooling layers, each of them dividing the output size of the filter stage by the chosen pooling kernel width. As a result, adding pooling layers *reduces* the input size of the classification stage, which in returns reduces the number of parameters of the network (as most parameters do lie in the classification stage).

The best performance for the raw experiment on the cross-validation set was found with: 270 ms of context, 10, 5 and 9 frames kernel width, 10, 1 and 1 frames shift, 90 filters, 500 hidden units and 3 pooling width. For the MFCC experiment, 30 frames (290 ms) context, 39, 5 and 7 kernel width, 80 filters and 500 hidden units were found the same way. The MFCC-based networks had no pooling layer. We found pooling operations were decreasing the performance with these features, while they are crucial for raw signal input experiments (see Section 6.1). This is not surprising, as MFCCs are sufficiently engineered to work well with simple network classifiers.

As a comparison, we also investigate traditional single hidden layer MLP-based approach [20]. Again, early stopping on the cross-validation set was used to determine the optimal number of nodes (500 nodes were found). The experiments were implemented using the *torch7* toolbox [21].

Table 1: Network hyper-parameters

| Parameter | Range |
|---|---|
| Input window size (ms) | 100-700 |
| Kernel width ($kW$) | 1-9 |
| Number of filters per kernel ($d_{out}$) | 10-90 |
| Number of hidden units in the class. stage | 100-1500 |

### 4.4. Decoding

We used the simple CRF approach described in Section 3.2 as decoding algorithm, with no duration constraints. We also report experimental results with a standard HMM decoder, with constrained duration of 3 states, and considering all phoneme equally probable.

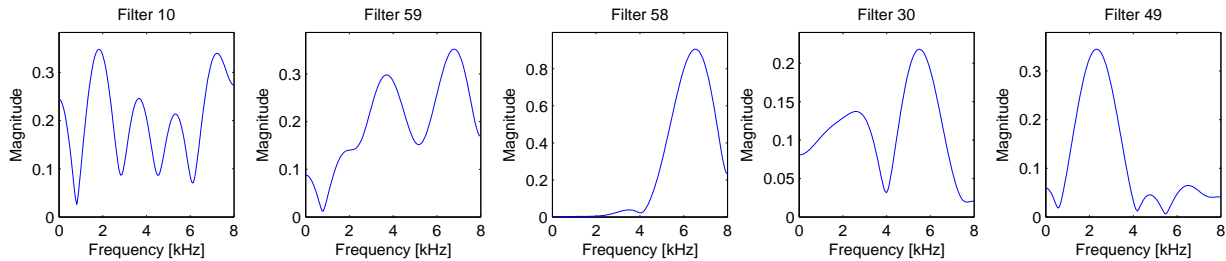

Figure 4: *Frequency responses of filters learned in the first convolutional layer.*

## 5. Results

We propose to evaluate the network capacity to estimate conditional probabilities by a phoneme sequence recognition experiment on the TIMIT database. The results are presented in Table 2, in term of phoneme accuracy for the different features and decoding scheme, along with the number of parameters. Using raw speech, the CNN architecture slightly outperforms the baseline, and the CRF approach increases the accuracy compared to the HMM approach. Using MFCC features with the CNN architecture yields similar performance as the raw features. The baseline accuracy is consistent with other works, although a bit lower, certainly due to the absence of supplementary processing, like speaker-level mean variance normalization in [22].

Table 2: Phoneme recognition accuracy on the core test set of TIMIT corpus.

| Features | Arch. | Decoding | Num. param. | Test acc. |
| --- | --- | --- | --- | --- |
| MFCC | MLP | HMM | 196'040 | 66.65 |
| Raw | MLP | HMM | 740'540 | 38.91 |
| Raw | CNN | HMM | 720'110 | 67.88 |
| Raw | CNN | CRF |  | 69.47 |
| MFCC | CNN | HMM | 860'700 | 70.52 |
| MFCC | CNN | CRF |  | 71.80 |

## 6. Analysis

### 6.1. Advantage of max-pooling layers

We varied the number of pooling layers, to evaluate their contribution in the overall performance of the architecture. The other hyper-parameters were tuned such that the same input window size was kept for each architecture. The output dimension of each convolution were also tuned for each case (to reduce over-fitting due to a too large number of parameters). The phoneme accuracy of each architecture is reported in Table 3, using raw features and HMM decoding, along with the number of parameters of the network. Clearly, adding max-pooling layer improves the system performance while providing an easy way to reduce the number of parameters (see Section 4.3).

### 6.2. Filters trained in the first layer

Figure 4 presents the frequency response of five randomly chosen filters[2]. Clearly, each filter learned by the network responds to different frequency bands of the input raw signal. These filters could be seen as matching filters. In a future work, we will investigate the relationship between the filters learned and the task at hand.

[2]Responses from all filters can be found at http://ronan.collobert.com/pub/extra/2013-is-cnn/filter-responses.pdf

Table 3: Max-pooling (MP) layers contribution

| Number of MP layers | Network parameters | Test Accuracy |
| --- | --- | --- |
| 3 | 303'460 | 67.60 |
| 2 | 380'660 | 67.18 |
| 1 | 507'860 | 67.14 |
| 0 | 593'460 | 64.96 |

## 7. Discussion

Over raw speech, the CNN architecture shows a great improvement compared to the single layer MLP architecture, confirming that convolution-based architectures are better suited for temporal signals. Moreover, it slightly outperforms the baseline, with almost no pre-processing on the data. These results suggest that deep architecture can learn efficient features and more importantly, that it is possible to achieve similar performances than complex hand-crafted features.

When comparing MFCC and raw signal as input for the CNN, MFCC seems to work slightly better. This aspect needs to be further investigated in the context of large database and using deeper architectures [11], where the slight advantage of MFCC might collapse.

When adding a decoder, the CRF approach seems to work better than the generative HMM approach. A plausible explanation is that the CRF learns a bigram language model over the phonemes. Also, in this work the CRF is optimized independently from the CNN, but joint training of the two models is in fact possible [23], and might lead to better performances.

## 8. Conclusions

In this paper, we proposed to use convolutional neural networks to estimate phoneme class probabilities. Our system is able to learn features by taking raw speech data as input and outperforms baseline systems. Moreover, using MFCC feature as input yields comparable performances. For future work, we plan to evaluate the robustness of our architecture with studies in noisy conditions. Secondly, as this work was intended as a first step for an end-to-end trained system, we plan to develop such a system applying the Graph Transformer Networks [23] approach. From there, we aim to develop more specific applications, such as Spoken Term Detection.

# 9. References

[1] Y. LeCun, L. Bottou, Y. Bengio, and P. Haffner, "Gradient-based learning applied to document recognition," *Proceedings of the IEEE*, vol. 86, no. 11, pp. 2278–2324, 1998.

[2] R. Collobert, J. Weston, L. Bottou, M. Karlen, K. Kavukcuoglu, and P. Kuksa, "Natural language processing (almost) from scratch," *The Journal of Machine Learning Research*, vol. 12, pp. 2493–2537, 2011.

[3] A. Waibel, T. Hanazawa, G. Hinton, K. Shikano, and K. Lang, "Phoneme recognition using time-delay neural networks," *Acoustics, Speech and Signal Processing, IEEE Transactions on*, vol. 37, no. 3, pp. 328–339, mar 1989.

[4] Y. LeCun, "Generalization and network design strategies," in *Connectionism in Perspective*, R. Pfeifer, Z. Schreter, F. Fogelman, and L. Steels, Eds. Zurich, Switzerland: Elsevier, 1989.

[5] A. Poritz, "Linear predictive hidden markov models and the speech signal," in *Acoustics, Speech, and Signal Processing, IEEE International Conference on ICASSP '82.*, vol. 7, May 1982, pp. 1291–1294.

[6] Y. Ephraim and W. J. J. Roberts, "Revisiting autoregressive hidden markov modeling of speech signals," *IEEE Signal Processing Letters*, vol. 12, no. 2, pp. 166–169, Feb. 2005.

[7] B. Mesot and D. Barber, "Switching linear dynamical systems for noise robust speech recognition," *IEEE Transactions on Audio, Speech, and Language Processing*, vol. 15, no. 6, pp. 1850–1858, Aug. 2008.

[8] H. Sheikhzadeh and L. Deng, "Waveform-based speech recognition using hidden filter models: Parameter selection and sensitivity to power normalization," *Speech and Audio Processing, IEEE Transactions on*, vol. 2, no. 1, p. 8089, 1994.

[9] J. Yousafzai, Z. Cvetkovic, and P. Sollich, "Tuning support vector machines for robust phoneme classification with acoustic waveforms," in *INTERSPEECH*, 2009, pp. 2391–2394.

[10] A. Mohamed, G. Dahl, and G. Hinton, "Acoustic modeling using deep belief networks," *Audio, Speech, and Language Processing, IEEE Transactions on*, vol. 20, no. 1, pp. 14–22, jan. 2012.

[11] O. Abdel-Hamid, A.-r. Mohamed, H. Jiang, and G. Penn, "Applying convolutional neural networks concepts to hybrid NN-HMM model for speech recognition," in *Acoustics, Speech and Signal Processing (ICASSP), 2012 IEEE International Conference on*, 2012, pp. 4277–4280.

[12] G. E. Dahl, D. Yu, L. Deng, and A. Acero, "Context-dependent pre-trained deep neural networks for large-vocabulary speech recognition," *Audio, Speech, and Language Processing, IEEE Transactions on*, vol. 20, no. 1, p. 3042, 2012.

[13] G. Hinton, L. Deng, D. Yu, G. E. Dahl, A.-r. Mohamed, N. Jaitly, A. Senior, V. Vanhoucke, P. Nguyen, and T. N. Sainath, "Deep neural networks for acoustic modeling in speech recognition: the shared views of four research groups," *Signal Processing Magazine, IEEE*, vol. 29, no. 6, p. 8297, 2012.

[14] Y. LeCun, F. J. Huang, and L. Bottou, "Learning methods for generic object recognition with invariance to pose and lighting," in *Proceedings of the IEEE Computer Society Conference on Computer Vision and Pattern Recognition*, vol. 2, 2004, pp. II–97.

[15] A. Krizhevsky, I. Sutskever, and G. Hinton, "Imagenet classification with deep convolutional neural networks," in *Advances in Neural Information Processing Systems 25*, 2012, pp. 1106–1114.

[16] J. Bridle, "Probabilistic interpretation of feedforward classification network outputs, with relationships to statistical pattern recognition," in *Neuro-computing: Algorithms, Architectures and Applications*, NATO ASI series ed., F. Fogelman Soulié and J. Hérault, Eds., 1990, pp. 227–236.

[17] L. Bottou, "Stochastic gradient learning in neural networks," in *Proceedings of Neuro-Nmes 91*. Nimes, France: EC2, 1991.

[18] K. F. Lee and H. W. Hon, "Speaker-independent phone recognition using hidden markov models," *IEEE Transactions on Acoustics, Speech and Signal Processing*, vol. 37, no. 11, pp. 1641–1648, 1989.

[19] S. Young, G. Evermann, D. Kershaw, G. Moore, J. Odell, D. Ollason, V. Valtchev, and P. Woodland, "The htk book," *Cambridge University Engineering Department*, vol. 3, 2002.

[20] N. Morgan and H. Bourlard, "Continuous speech recognition," *Signal Processing Magazine, IEEE*, vol. 12, no. 3, pp. 24–42, May 1995.

[21] R. Collobert, K. Kavukcuoglu, and C. Farabet, "Torch7: A matlab-like environment for machine learning," in *BigLearn, NIPS Workshop*, 2011.

[22] S. Thomas, S. Ganapathy, and H. Hermansky, "Phoneme recognition using spectral envelope and modulation frequency features," in *Acoustics, Speech and Signal Processing, 2009. ICASSP 2009. IEEE International Conference on*, 2009, pp. 4453–4456.

[23] L. Bottou, Y. Bengio, and Y. LeCun, "Global training of document processing systems using graph transformer networks." in *In Proc. of Computer Vision and Pattern Recognition*. Puerto-Rico., 1997, pp. 490–494.